\documentclass[sigconf]{acmart}

\usepackage{makecell}
\usepackage{algorithm}
\usepackage{enumitem}

\AtBeginDocument{%
  \providecommand\BibTeX{{%
    \normalfont B\kern-0.5em{\scshape i\kern-0.25em b}\kern-0.8em\TeX}}}

\copyrightyear{2023} 
\acmYear{2023} 
\setcopyright{acmlicensed}\acmConference[K-CAP '23]{Knowledge Capture
Conference 2023}{December 5--7, 2023}{Pensacola, FL, USA}
\acmBooktitle{Knowledge Capture Conference 2023 (K-CAP '23), December
5--7, 2023, Pensacola, FL, USA}
\acmPrice{15.00}
\acmDOI{10.1145/3587259.3627566}
\acmISBN{979-8-4007-0141-2/23/12}

\begin{document}

\title{Capturing Pertinent Symbolic Features for Enhanced Content-Based Misinformation Detection}

\author{Flavio Merenda}
\affiliation{%
  \institution{expert.ai}
  \streetaddress{}
  \city{Madrid}
  \state{}
  \country{Spain}
  \postcode{}
}
\email{fmerenda@expert.ai}
\orcid{0009-0001-4136-5415}

\author{José Manuel Gómez Pérez}
\affiliation{%
  \institution{expert.ai}
  \streetaddress{}
  \city{Madrid}
  \state{}
  \country{Spain}
  \postcode{}
}
\email{jmgomez@expert.ai}
\orcid{0000-0002-5491-6431}

\renewcommand{\shortauthors}{Merenda and Gómez-Pérez}

\begin{abstract}
  Preventing the spread of misinformation is challenging. The detection of misleading content presents a significant hurdle due to its extreme linguistic and domain variability. Content-based models have managed to identify deceptive language by learning representations from textual data such as social media posts and web articles. However, aggregating representative samples of this heterogeneous phenomenon and implementing effective real-world applications is still elusive. Based on analytical work on the language of misinformation, this paper analyzes the linguistic attributes that characterize this phenomenon and how representative of such features some of the most popular misinformation datasets are. We demonstrate that the appropriate use of pertinent symbolic knowledge in combination with neural language models is helpful in detecting misleading content. Our results achieve state-of-the-art performance in misinformation datasets across the board, showing that our approach offers a valid and robust alternative to multi-task transfer learning without requiring any additional training data. Furthermore, our results show evidence that structured knowledge can provide the extra boost required to address a complex and unpredictable real-world problem like misinformation detection, not only in terms of accuracy but also time efficiency and resource utilization.
\end{abstract}




%   \caption{}
%   \Description{}

\maketitle

\section{Introduction}
Online misinformation is one of the biggest challenges societies are facing nowadays, even though the story of this phenomenon is as old as the world \cite{ecker2022psychological}. Online communication and social media amplify the circulation of false information to a scale and speed never seen in history \cite{allcott2019trends}. Moreover, misinformation spreads six times faster than factual information exposing internet users to the construction of false beliefs, difficult to contrast and eradicate \cite{vosoughi2018spread}. 
Over the years, researchers have joined efforts to implement models that attempt to detect deceptive content and thereby mitigate and reduce the spread of online misinformation \cite{guo2019future}. 
Despite the remarkable capabilities unveiled by recent advancements in natural language processing for the classification and analysis of written texts \cite{li2022survey}, this phenomenon remains intricate and far from resolved \cite{pendyala2023spectral}. The challenges of heterogeneity \cite{carrasco2022fingerprints} and cross-modality \cite{micallef2022cross} make it exceedingly difficult to observe this phenomenon at the necessary volume and variety required to curate annotated datasets essential for training effective and generalizable models through supervised approaches. 
Nevertheless, the progress in generative large language models such as GPT-3 \cite{brown2020language} and PaLM \cite{narang2022pathways}, disclosed alarming scenarios in the automatic generation of misleading content, becoming a possible undesirable tool in the hands of mala fide actors \cite{wahl2023news}.
The research questions we address in this paper are the following:
\begin{itemize}
    \item RQ1: What are the main attributes of the datasets currently adopted by the community of researchers to characterize the misinformation problem?
    %Are training datasets used in the community for online content-based misinformation detection representative of the phenomenon characterized by social science studies?
    \item RQ2: Taking into account existing work on linguistic and psychological drivers of misinformation, are the linguistic attributes they propose predictive?%, and what are the patterns that emerge from the different training datasets?
    \item RQ3: Considering the heterogeneity of this phenomenon, could structured features proposed by social science studies, as captured by a collection of pre-existing symbolic models, enhance the development of more robust content-based misinformation detection models?
\end{itemize}

To address these questions, we first analyze datasets used nowadays in the community and collected with the intention of representing the diversity of the misinformation. We extract specific features that are supposed to characterize misinforming texts and confirm the representativeness of such data.
Next, using feature selection methods, we demonstrate the predictability of these features and briefly analyze the patterns that emerge across the different training data sources.
Finally, we experiment with combining these features with neural language models to explore the utility of these resources in building machine learning models for content-based misinformation detection. We hypothesize that incorporating these features can improve the models' robustness in terms of generalizability and the capability to withstand domain shifting. 

\section{Related Work}
We review the related work on misinformation concerning our main contributions to (i) the characterization of content spread by this phenomenon and (ii) the models developed to detect such content. In this work, we adopt the definition of misinformation presented in \cite{wu2019misinformation}, or rather, {\itshape an umbrella
term to include all false or inaccurate information that is
spread} online, such as {\itshape rumor}, {\itshape clickbait} or {\itshape fake news}, among others \cite{wu2019misinformation, carrasco2022fingerprints, lee-etal-2021-unifying}, intentionally or unintentionally propagated. Moreover, we only consider content-based classification models, which rely exclusively on textual data from various misleading online sources, such as web articles or social media posts, supporting content {\itshape pre-bunking} \cite{ecker2022psychological}. We do not consider models that leverage additional sources such {\itshape social-data based} methods \cite{alam-etal-2022-survey} nor models that make use of external sources for content {\itshape debunking} such as evidence-based methods \cite{denaux2020linked}.

\subsection{Analysis of Misleading Contents}
Several works over the years have been investigating the language of misleading content spread by misinformation.
Recent social studies analyzing linguistic and psychological drivers of misinformation identify the relevance of specific linguistic features in the characterization of misleading content. In \cite{carrasco2022fingerprints}, the author investigates the distinctions between reliable sources and untrustworthy ones concerning cognitive effort and emotional appeal, highlighting the importance of text attributes such as its readability, sentiment, or social identity among others.
Authors in \cite{ecker2022psychological} emphasize the importance of emotions in the formation of false beliefs, while previous research \cite{hu2014social}, similarly, reveals the importance of sentiment information as indicators of deceptive content. A higher level of abstraction is considered in \cite{rubin2015truth}, where the authors identify systematic differences between deceptive and truthful content in rhetorical structures. Analogously, thematic content analysis is investigated in \cite{golbeck2018fake}, in which a set of narratives and rhetorical patterns define intents spread by such content.

Various studies delve into more granular levels of linguistic analysis. In \cite{grieve2023language}, a framework is introduced, which employs grammatical patterns to distinguish between authentic and deceptive news. Similarly, in \cite{rashkin2017truth}, the authors identify lexical indicators for this purpose. A work that tries to combine lexical, psychological, as well as more complex structural features is presented in \cite{horne2017just}, in which the authors combine different levels of analysis to extract common patterns in fake news content.

Valuable insights arise from all of these works. Typically, misleading content displays negative sentiment, employs emotional appeals, and incorporates first or second-person references. It adheres to particular rhetorical structures and narratives. These characteristics contribute to the content's accessibility and resonance with the audience.
In our study, we conduct a comprehensive layered linguistic analysis of misleading content. This analysis encompasses both fine-grained and coarse-grained traits, spanning from lexical to discourse attributes, including all language aspects identified in related research \cite{rashkin2017truth, carrasco2022fingerprints, ecker2022psychological, hu2014social, horne2017just, rubin2015truth}.

\subsection{Content-based Misinformation Detection}

The use of linguistic features to classify misleading content has been experimented extensively with traditional machine learning algorithms. In \cite{qazvinian2011rumor} a {\itshape Bayes} classifier has been used to detect misinforming rumors in microblogs, while {\itshape Bayes}, {\itshape Decision tree} and {\itshape SVM} classifiers have been implemented in \cite{castillo2011information} to assess the credibility of tweets. The importance of features related to upper layers of language analysis has been explored in \cite{rubin2015truth}, in which the authors leverage the vectorization of rhetoric information to cluster a dataset of personal stories, divided between truthful and deceptive.

The use of neural network models in combination with symbolic features has also been investigated across various studies. LSTM models to detect misinforming articles have been explored in different works \cite{rashkin2017truth, giachanou2019leveraging}. These models leverage the combination of word embeddings with linguistic features, such as the {\itshape EmoCred} system, which experiments with the use of lexical resources and attention-based methods taking advantage of emotional signals from texts. A work that extends the use of {\itshape EmoCred} with the transformers \cite{vaswani2017attention} is presented in \cite{kelk2022automatic}.
The exploration of model generalizability across various misinformation data sources has been pursued by employing transformers in multi-task learning \cite{panda2022improving} and transfer learning \cite{lee-etal-2021-unifying} contexts. 

Previous research provides robust evidence supporting the efficacy of linguistic features extracted from multiple language layers in predicting deceptive content. Some studies underscore the benefits of combining these resources with advanced language representations, such as neural embeddings, proposing effective integration methodologies.
Nonetheless, a systematic investigation that leverages all these features concurrently and offers comprehensive validation across diverse misinformation sources is currently lacking. In this work, we build on the findings of previous studies and conduct additional experiments to address the remaining challenges.

\begin{table*}
  \Description{This table summarizes the four misinformation datasets utilized in the paper's experiments for training language models under a complete fine-tuning scenario. It includes details such as the dataset name and source, the specific task, granularity, labels, their distribution, and the dataset size.}
  \caption{Summary of the 4 misinformation datasets used to train model in fully finetuning scenario.}
  \begin{tabular}{cccccc}
    \toprule
    Dataset Name & Task & Granularity & Labels (Positive/Negative) & Dataset Size & Positive Class Size \\
    \midrule
    BASIL \cite{fan2019plain} & NewsBias & sentence &  contains-bias/no-bias &  7,959 &  1,624 \\
    Webis \cite{potthast2017stylometric} & FakeNews & article & fake/true & 1,604 & 355 \\
    PHEME \cite{zubiaga2016analysing} & Rumor & tweet & True/False & 1,685 & 1,058 \\
    Clickbait \cite{potthast2018webis} & Clickbait & headline & is-clickbait/not-clickbait & 19,038 &  4,318 \\
  \bottomrule
  \label{table:1}
\end{tabular}
\end{table*}

\begin{table*}
  \Description{This table summarizes the five misinformation datasets utilized in the paper's experiments for training language models under fewshot learning scenario. It includes details such as the dataset name and source, the specific task, granularity, labels, their distribution, and the dataset size.}
  \caption{Summary of the 4 misinformation datasets used to train model in fewshot learning scenario.}
  \resizebox{\textwidth}{!}{\begin{tabular}{cccccc}
    \toprule
    Dataset Name & Task & Granularity & Labels (Positive/Negative) & Dataset Size & Positive Class Size \\
    \midrule
    PropagandaTC \cite{da2019fine} & Propaganda & article &  has\_propaganda/no\_propaganda &  1,594 &  816 \\
    PolitiFact \cite{shu2019beyond} & Fake News Article & article & fake/real & 202 & 91 \\
    BuzzFeed \cite{shu2019beyond} & Fake News Title & headline & fake/real & 170 & 80 \\
    CovidTwitterQ1 \cite{alam2020fighting} & Covid Check-worthy Twitter & tweet & yes/no & 504 &  305 \\
    CovidTwitterQ2 \cite{alam2020fighting} & Covid False Twitter Claim & tweet & contains\_false/no\_false & 260 & 37 \\
  \bottomrule
  \label{table:2}
  \end{tabular}}
\end{table*}

\section{Resources}
In this section, we introduce the data and linguistic resources to be employed in this study, considering the insights from the related work discussed in Section 2.

\subsection{Data}
We select datasets from \cite{lee-etal-2021-unifying}, which consolidates various forms of misinformation, domains, and text structures, providing a comprehensive representation of the phenomenon. This list employs 9 datasets, divided into 2 main groups, that have undergone manual annotation in contrast to distantly supervised data used in other research \cite{rashkin2017truth, carrasco2022fingerprints}. The first group, summarized in Table \ref{table:1}, has been used by the authors to fully finetune a RoBERTa large model \cite{liu1907roberta} in a multitask learning scenario \cite{crawshaw2020multi}, where each task refers to a different type of misinformation. The second group, summarized in Table \ref{table:2}, has been used by the authors to evaluate model generalizability on new unknown tasks in a few-shot learning scenario \cite{song2205comprehensive}.

\subsection{Symbolic Models}
Based on the analysis performed in section 2.1 on the formal and semantic aspects of the language of misinformation suggested by prior studies, we select a collection of symbolic models that capture such linguistic attributes. Below, we introduce the selected models, organized from lower to higher linguistic layers of analysis that include formal, semantic, and discourse analysis.

\begin{description}
    \item[Writeprint]\footnote{\url{https://docs.expert.ai/nlapi/latest/reference/output/detection/writeprint/}} This model is capable of extracting basic linguistic attributes, such as stylometric traits or text statistics, and calculating widely used readability indexes that are employed to assess the cognitive effort required for writing and comprehending texts.

    \item[Sentiment analysis]\footnote{\url{https://docs.expert.ai/nlapi/latest/guide/sentiment-analysis/}} This is a type of document analysis that determines how positive or negative the tone of the text is.

    \item[Emotional traits]\footnote{\url{https://docs.expert.ai/nlapi/latest/guide/classification/emotional-traits/}} Classify documents in terms of the feelings expressed in the text. They can recognize different emotional traits such as joy, surprise, irritation, etc.

    \item[Behavioral traits]\footnote{\url{https://docs.expert.ai/nlapi/latest/guide/classification/behavioral-traits/}} Identify references to personality traits mentioned in the text, such as curiosity, honesty, negativity, etc.

    \item[Hate speech]\footnote{\url{https://docs.expert.ai/nlapi/latest/reference/output/detection/hate-speech/}} This model is designed to both extract the single instances of offensive and violent language and categorize each instance according to different hate speech categories.

    \item[Radicalization Narratives] \cite{denaux2019textual} Helps to capture wider discourse intents and strategic radicalization narratives that can be exploited to promote radical ideologies by any radical group.

\end{description}

To encompass the insights that have arisen from related work analysis, we apply the writeprint model to extract fundamental language attributes, including text statistics and stylometric traits, and to compute readability indices to assess the necessary cognitive effort required to process texts. To capture the emotional appeals and negative sentiment usually present in misinforming content, we utilize emotional traits, behavioral traits, hate speech, and sentiment analysis models. Finally, we utilize the radicalization narratives model to detect broader discourse intentions influenced by structural and rhetorical aspects of texts.

We use a collection of expert.ai's symbolic, rule-based models\footnote{\url{https://www.expert.ai/blog/symbolic-approach-nlp-models/}} to extract features that can be used to enhance machine learning classification algorithms. We opt for off-the-shelf models for linguistic analysis,\footnote{\url{https://docs.expert.ai/nlapi/latest/guide/linguistic-analysis/}} classification,\footnote{\url{https://docs.expert.ai/nlapi/latest/guide/classification/}} and information detection,\footnote{\url{https://docs.expert.ai/nlapi/latest/guide/detection/}} designed to encompass both fine-grained as well as course-grained features. 
Models are designed to assign weighted scores that quantify the relevance of such extracted features.
The expert.ai API\footnote{\url{https://github.com/therealexpertai/nlapi-python}} offers free easy access to all symbolic models, and they are available for quick testing using the online demo\footnote{\url{https://try.expert.ai/}}.

\section{Feature selection}
In this work, with the aim of capturing the optimal subset of features that exhibit predictive capability, we make use of feature selection methods \cite{venkatesh2019review}.
Typically, this process uses statistical methods to estimate relationships between input variables and the target variable, aiding in selecting features with stronger associations. For our analysis, we employ the {\itshape univariate linear regression tests} leveraging {\itshape Pearson correlation} \cite{venkatesh2019review}, where each input variable is tested separately as a single regressor.

let $D$ be a dataset containing $N$ text instances, let ${M}$ be a collection of $K$ different symbolic models where each model $M_K$ assigns a set of features $V$ to each instance in $D$. $V$ can be represented as a set of features $\{x_1, x_2, \ldots, x_i\}$, where $x_i$ represents an input variable of vector $V_N^{(K)}$.
The process comprises the calculation of the {\itshape Person correlation coefficient} ($r$) subsequently converted to {\itshape F-statistics} ($F_{stats}$) and {\itshape p-value} ($p$) as follows:

\begin{equation}
  r =
  \frac{ \sum_{i=1}^{n}(x_i-\bar{x})(y_i-\bar{y}) }{%
        \sqrt{\sum_{i=1}^{n}(x_i-\bar{x})^2}\sqrt{\sum_{i=1}^{n}(y_i-\bar{y})^2}}
\end{equation}
\begin{equation}
F_{stats} = \frac{r^2}{1 - r^2} \times \frac{n - 2}{1}
\end{equation}
\begin{equation}
p = 1 - \text{sf}_{F_{stats}}(F_{stats}, 1, n - 2)
\end{equation}
where $\text{sf}_{F_{stats}}$ is the survival function of the F-distribution.
Finally, let be $F^{(K)}$ a set of features assigned by model $M_K$. To select the predictive features for each model-task pair we calculate the reduced set of feature $F'^{(K)}$, obtained by considering only those features $f_j^{(K)}$, the $j$-th feature assigned by the $K$-th model where $f \in F^{(K)}$, for which the p-value $p_j^{(K)}$ from the test is less than or equal to $\alpha$, a predetermined significance level that we set to 0.05:
\begin{equation}
F'^{(K)} = \{f_j^{(K)} \mid p_j^{(K)} \leq \alpha\}
\end{equation}

\section{Proposed Model}
We employ RoBERTa large for our misinformation content-based classifier, as in \cite{lee-etal-2021-unifying}. We use adapters fine-tuning with the Pfeiffer architecture \cite{pfeiffer2020AdapterHub} and adapter drop method \cite{ruckle2020adapterdrop} to efficiently manage multiple dataset training, avoiding full model fine-tuning. We integrate insights from symbolic models presented in section 3.2 using a knowledge combination mechanism inspired by previous works \cite{giachanou2019leveraging, kelk2022automatic}.
We create a unified feature vector representation incorporating multiple models' information as follows. Given a set $S$ of text samples for a misinformation task and a set $M$ representing different models used for feature extraction, we compute vectors $v_{s,M_i}$ for each text sample $s \in S$ using model $M_i$. Starting with $v_{s,M_i}$, we derive a condensed feature vector $v_{s,M_i}'$ by isolating statistically significant features, as detailed in section 4. Concatenating these reduced feature vectors yields a unique representation, $v_{s,concat}'$. that contains the information from all the different models for sample $s$:
\begin{equation}
v_{s,concat}' = v_{s,M_1}' \oplus v_{s,M_2}' \oplus \ldots \oplus v_{s,M_n}'
\end{equation}
where $\oplus$ stands for concatenation. This vector is then fed to the classification head along with the RoBERTa embedding previously passed through a mean pooling layer. To weigh contributions coming from both representations, the feature vector and the transformer embedding are passed through normalized linear layers. These layers are subsequently concatenated, followed by the application of a Softmax function to calculate the prominence of the two representations:

\begin{displaymath}
\hat{f}_j = LayerNorm(W_F,  f_j + b_f)
\end{displaymath}
\begin{equation}
\hat{e}_j = LayerNorm(W_E,  e_j + b_e)
\end{equation}
\begin{displaymath}
softmax(\hat{f}_j \oplus \hat{e}_j)
\end{displaymath}

\begin{figure}[tb]
  \Description{This image depicts the architecture of the hybrid model proposed in this work. It illustrates the various layers of the network, including input representations, their weighting, the concatenated final representation, and the classification output.}
  \centering
  \includegraphics[width=\linewidth]{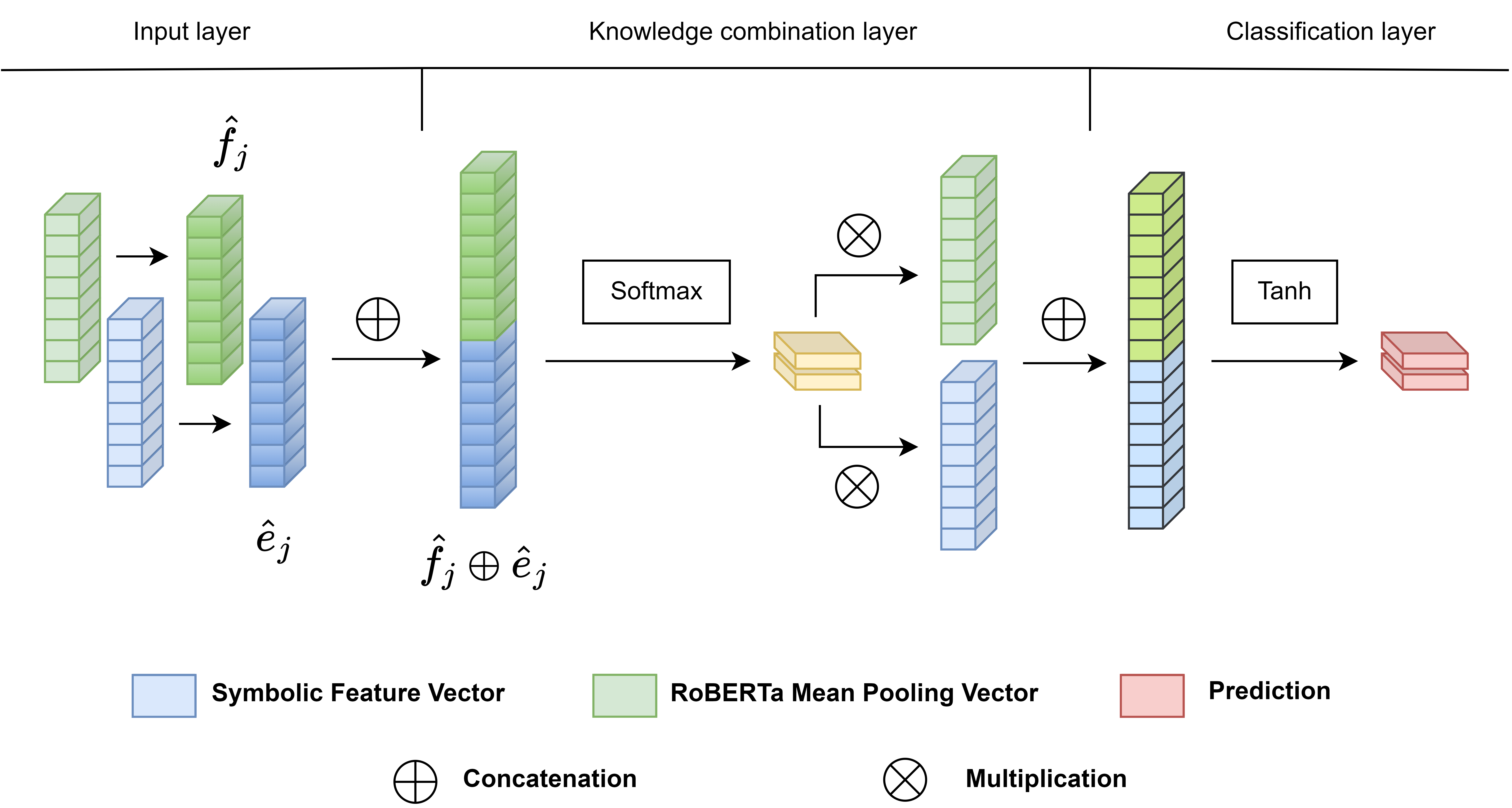}
  \caption{Architecture of the proposed model.}
  \label{fig:architecture}
\end{figure}

Element-wise product is successively computed between those weighted values and the original vectors before moving to a two-classification layer. A detailed overview of the model architecture is presented in Figure \ref{fig:architecture}.

\section{Experiments}
In this section, we present the experimental design of our study and the results of our analyses, which address the research questions that we posed in the introduction.\footnote{A repository containing both resources and the experimental code is accessible at \url{https://github.com/expertailab/Capturing-Pertinent-Symbolic-Features-for-Enhanced-Content-Based-Misinformation-Detection}}

\begin{table*}
  \Description{Frequency values of the top 3 detected classes within the misinformation datasets by each expert.ai symbolic model. The table clearly demonstrates that the features highlighted by social science studies are present in these datasets.}
  \caption{Frequency values of the top 3 detected classes within the misinformation datasets by each expert.ai symbolic model.}
  \begin{tabular}{ccccc}
    \toprule
    
    Dataset Name & Emotional Traits & Behavioural Traits & Hate Speech & Radicalization Narratives \\
    
    \midrule
    
    BASIL & \thead{Hatred (9.63\%) \\ Love (8.03\%) \\ Worry (8.02\%)} & \thead{Initiative (10.96\%) \\ Violence (7.86\%) \\ Rejection (7.69\%)} & \thead{Threat and Violence (45.8\%) \\ Ableism (13.94\%) \\ Religious Hatred (10.08\%)} & \thead{Legitimacy of ideology (97.05\%) \\ Homophily (1.2\%) \\ Group's achievements (0.85\%) } \\
    
    Webis & \thead{Hatred (10.7\%) \\ Fear (8.3\%) \\ Anger (7.2\%)} & \thead{Violence (14.17\%) \\ Extremism (6.76\%) \\ Initiative (6.51\%)} & \thead{Threat and Violence (33.17\%) \\ Racism (21.07\%) \\ Personal Insult (15.57\%)} & \thead{Legitimacy of ideology (96.24\%) \\ Group's achievements (1.19\%) \\ Homophily (0.99\%)} \\
    
    PHEME & \thead{Fear (27.48\%) \\ Anger (11.97\%) \\ Hatred (10.12\%)} & \thead{Violence (54.54\%) \\ Extremism (10.77\%) \\ Unlawfulness (6.1\%)} & \thead{Threat and Violence (47.59\%) \\ Racism (33.95\%) \\ Ableism (11.07\%)} & \thead{Legitimacy of ideology (75.0\%) \\ Homophily (13.89\%) \\ Group's achievements (11.11\%)} \\
    
    Clickbait & \thead{Love (10.35\%) \\ Fear (8.14\%) \\ Anger (7.78\%)} & \thead{Violence (14.71\%) \\ Unlawfulness (8.92\%) \\ Initiative (8.68\%)} & \thead{Threat and Violence (39.26\%) \\ Racism (26.16\%) \\ Personal Insult (10.07\%)} & \thead{Legitimacy of ideology (82.64\%) \\ Homophily (9.99\%) \\ Group's achievements (5.3\%)} \\
    
    PropagandaTC & \thead{Hatred (15.88\%) \\ Fear (11.98\%) \\ Well-Being (8.05\%)} & \thead{Violence (18.01\%) \\ Initiative (9.85\%) \\ Unlawfulness (7.33\%)} & \thead{Threat and Violence (28.56\%) \\ Racism (19.01\%) \\ Religious Hatred (14.18\%)} & \thead{Legitimacy of ideology (83.64\%) \\ Group's achievements (8.39\%) \\ Homophily (3.5\%)} \\
    
    PolitiFact & \thead{Anger (12.0\%) \\ Hatred (8.67\%) \\ Love (8.6\%)} & \thead{Violence (13.06\%) \\ Initiative (10.14\%) \\ Rejection (5.32\%)} & \thead{Threat and Violence (37.93\%) \\ Racism (22.06\%) \\ Personal Insult (14.61\%)} & \thead{Legitimacy of ideology (90.71\%) \\ Group's achievements (3.42\%) \\ Homophily (2.1\%)} \\
    
    BuzzFeed  & \thead{Hatred (20.21\%) \\ Anger (10.45\%) \\ Disgust (10.45\%)}  & \thead{Extremism (20.5\%) \\ Unlawfulness (10.93\%) \\ Discrimination (9.68\%)} & \thead{Personal Insult (50.0\%) \\ Racism (24.1\%) \\ Threat and Violence (24.1\%)} & \thead{Legitimacy of ideology (100.0\%) \\ - \\ -} \\
    
    CovidTwitterQ1 & \thead{Guilt (20.06\%) \\ Anger (17.38\%) \\ Hatred (9.74\%)}  & \thead{Isolation (11.46\%) \\ Initiative (7.52\%) \\ Unawareness (5.96\%)}  & \thead{Threat and Violence (29.51\%) \\ Personal Insult (27.66\%) \\ Racism (16.38\%)}  & \thead{Legitimacy of ideology (75.0\%) \\ Homophily (15.62\%) \\ Promote group ideology (9.38\%)} \\
    
    CovidTwitterQ2 & \thead{Guilt (32.84\%) \\ Fear (9.66\%) \\ Hatred (9.14\%)} & \thead{Isolation (13.91\%) \\ Initiative (10.82\%) \\ Rejection (6.8\%)} & \thead{Threat and Violence (29.59\%) \\ Personal Insult (26.53\%) \\ Body Shaming (18.37\%)} & \thead{Legitimacy of ideology (70.59\%) \\ Homophily (29.41\%) \\ -} \\
    
  \bottomrule
  
  \label{table:3}
  \end{tabular}
\end{table*}

\begin{table*}
  \Description{Sentiment and Writeprint analysis performed by expert.ai symbolic models on the misinformation datasets. The results emphasize the prevailing negative sentiment throughout the datasets, as well as their straightforward linguistic structure.}
  \caption{Sentiment and Writeprint analysis performed by expert.ai symbolic models on the misinformation datasets.}
  \begin{tabular}{ccccccc}
    \toprule
    \multicolumn{1}{c}{} &
    \multicolumn{3}{c}{Sentiment} &
    \multicolumn{3}{c}{Writeprint}  \\
    \cmidrule(lr){2-4}
    \cmidrule(lr){5-7}
    \multicolumn{1}{l}{Dataset Name} &
    \multicolumn{1}{c}{Positivity} &
    \multicolumn{1}{c}{Negativity} &
    \multicolumn{1}{c}{Overall} &
    \multicolumn{1}{c}{Coleman-Liau} &
    \multicolumn{1}{c}{Gulpease} &
    \multicolumn{1}{c}{Automated Readability} \\
    \midrule
    BASIL & 2.1 & -4.4 & -2.3 & 10.3 & 58.1 & 13.2 \\
    Webis & 2.4 & -5.6 & -3.1 & 9.9 & 56.9 & 11.8 \\
    PHEME & 1.5 & -10.1 & -8.6 & 18.3 & 55.9 & 14.5 \\
    Clickbait & 4.1 & -7.9 & -3.7 & 9.8 & 69.4 & 7.3 \\
    PropagandaTC & 1.9 & -7.4 & -5.4 & 8.4 & 74.0 & 10.9 \\
    PolitiFact & 2.6 & -5.9 & -3.1 &  9.5 & 58.9 & 10.3 \\
    BuzzFeed  & 1.6 & -15.4 & -13.8 &  9.5 & 72.3 & 6.7 \\
    CovidTwitterQ1 & 2.3 & -8.8 & -6.4 & 11.4 & 63.7 & 10.7 \\
    CovidTwitterQ2 & 2.3 & -8.7 & -6.3 & 11.4 & 61.3 & 10.9 \\
  \bottomrule
  \label{table:4}
  \end{tabular}
\end{table*}

\begin{table*}
  \Description{The top 3 features with their corresponding p-values obtained through univariate linear regression within the misinformation tasks for each expert.ai symbolic model. P-values $\leq$ 0.05 indicate significance. The lower the value, the bigger the confidence. Not all the most frequent features are predictive for the reference tasks, but some are, and notably, less common features are predictive for specific datasets.}
  \caption{The top 3 features with their corresponding p-values obtained through univariate linear regression within the misinformation tasks for each expert.ai symbolic model. P-values $\leq$ 0.05 indicate significance. The lower the value, the bigger the confidence.}
  \resizebox{\textwidth}{!}{\begin{tabular}{ccccccc}
  
    \toprule
    
    Task & Emotioanl Traits & Behavioural Traits & Hate Speech & Radicalization Narratives & Sentiment & Writeprint \\
    
    \midrule
    
    NewsBias & \thead{Worry (0.0) \\ Shame (0.0) \\ Offence (0.0)} & \thead{Impoliteness (0.0) \\ Calmness (0.0) \\ Rejection (0.0)} & \thead{-\\ -\\ -} & \thead{Group's achievements (0.05) \\ -\\ -} & \thead{negativity (0.0) \\ overall (0.0) \\ positivity (0.0)} & \thead{Gulpease (0.0) \\ Automated Readability (0.0) \\ adjectivesPerSentence (0.0)} \\
    
    FakeNews & \thead{Worry (0.01) \\ Satisfaction (0.01) \\ Hatred (0.03)} & \thead{Initiative (0.0) \\ Progressiveness (0.0) \\ Competence (0.0)} & \thead{Religious Hatred (0.0) \\ Classism (0.01) \\ -} & \thead{Legitimacy of ideology (0.0) \\ Discredit enemy (0.01) \\ Attack as self-defense (0.01)} & \thead{negativity (0.0) \\ overall (0.0) \\ positivity (0.0)} & \thead{Coleman-Liau (0.0) \\ Gulpease (0.0) \\ Automated Readability (0.0)} \\
    
    Rumor & \thead{Surprise (0.0) \\ Disappointment (0.02) \\ Hatred (0.04)} & \thead{Bias (0.0) \\ Violence (0.0) \\ Humour (0.02)} & \thead{Racism (0.01) \\ -\\ -} & \thead{Homophily (0.01) \\ -\\ -} & \thead{negativity (0.0) \\ overall (0.0) \\ positivity (0.02)} & \thead{Coleman-Liau (0.0) \\ Gulpease (0.0) \\ Automated Readability (0.0)} \\
    
    Clickbait & \thead{Surprise (0.0) \\ Happiness (0.0) \\ Amusement (0.0)} & \thead{Seriousness (0.0) \\ Humour (0.0) \\ Pleasantness (0.0)} & \thead{Racism (0.0) \\ - \\ -} & \thead{Legitimacy of ideology (0.0) \\ Discredit enemy (0.05) \\ -} & \thead{negativity (0.0) \\ overall (0.0) \\ positivity (0.0)} & \thead{Coleman-Liau (0.0) \\ Gulpease (0.0) \\ Automated Readability (0.0)} \\
    
    Propaganda & \thead{Hatred (0.01) \\ Disgust (0.02) \\ Offence (0.05)} & \thead{Violence (0.0) \\ Extremism (0.0) \\ Disagreement (0.01)} & \thead{Religious Hatred (0.0) \\ -\\ -} & \thead{Legitimacy of ideology (0.0) \\ Discredit enemy (0.02) \\ -} & \thead{negativity (0.0) \\ overall (0.0) \\ -} & \thead{Coleman-Liau (0.0) \\ Gulpease (0.0) \\ Automated Readability (0.0)} \\
    
    Fake News Article & \thead{Surprise (0.0) \\ Hatred (0.04) \\ -} & \thead{Addiction (0.01) \\ Impoliteness (0.03) \\ Organization (0.03)} & \thead{Racism (0.01) \\ Sexism (0.03) \\ -} & \thead{Legitimacy of ideology (0.0) \\ -\\ -} & \thead{negativity (0.05) \\ -\\ -} & \thead{auxiliariesPerSentence (0.0) \\ commasPerSentence (0.0) \\ doubleQuotationMarksPerSentence (0.0)} \\
   
    Fake News Title & \thead{-\\ -\\ -} & \thead{-\\ -\\ -} & \thead{-\\ -\\ -} & \thead{-\\ -\\ -} & \thead{-\\ -\\ -} & \thead{colonsPerSentence (0.0) \\ exclamationMarksPerSentence (0.0) \\ sentences (0.0)} \\
    
    Covid Check-worthy Twitter & \thead{-\\ -\\ -} & \thead{Initiative (0.02) \\ Apprehension (0.04) \\ Emotionality (0.05)} & \thead{-\\ -\\ -} & \thead{-\\ -\\ -} & \thead{-\\ -\\ -} & \thead{Gulpease (0.0) \\ capitalFirstLetterSentences (0.0) \\ charactersPerSentence (0.0)} \\
   
    Covid False Twitter Claim & \thead{Anger (0.01) \\ Anxiety (0.01) \\ Worry (0.01) } & \thead{Sexuality (0.01) \\ Unawareness (0.02) \\ Discrimination (0.04)} & \thead{-\\ -\\ -} & \thead{-\\ -\\ -} & \thead{-\\ -\\ -} & \thead{tokens (0.0) \\ emoticonsPerSentence (0.01) \\ Automated Readability (0.03)} \\
    
  \bottomrule
  
  \label{table:5}
  \end{tabular}}
\end{table*}

\subsection{Characterization of Misinformation Datasets}

We answer RQ1 by investigating the characterization of the datasets introduced in section 3.1, Table \ref{table:1} and Table \ref{table:2}, and verifying their alignment with the representations outlined in social science studies, as discussed in section 2.1. This process offers an assessment of the data's representativeness and furnishes valuable insights into their similarities provided by the application of symbolic models discussed in section 3.2.
Given the same formalism presented in section 4, a dataset can be represented as:
\begin{equation}
D = \{(V_1^{(1)}, V_1^{(2)}, \ldots, V_1^{(K)}), \ldots, (V_N^{(1)}, V_N^{(2)}, \ldots, V_N^{(K)})\}
\end{equation}
where $V_N^{(K)}$ represents the feature vector assigned by model $M_K$ to each $N$-th instance of $D$ and previously described as a set of features $\{x_1, x_2, \ldots, x_i\}$. To understand the magnitude of each feature $x_i$ across the feature vectors $V_N^{(K)}$, we compute the average of such vectors assigned by model $M_K$ to each text $N$-th instance of $D$ as follows:
\begin{equation}
\bar{V}^{(K)} = \frac{1}{N}\sum_{j=1}^{N} V_j^{(K)}
\end{equation}

We provide the results of the analysis in Table \ref{table:3} and Table \ref{table:4}. Specifically, Table \ref{table:3} shows the top 3 feature values, or classes, captured from each $\bar{V}^{(K)}$. Across the datasets, the predominant classes consistently confirm the highly emotional nature of deceptive content, often characterized by negative and aggressive emotions. These emotions are well-represented by labels such as {\itshape Hatred}, {\itshape Anger}, {\itshape Violence}, {\itshape Extremism}, {\itshape Discrimination}, {\itshape Isolation}, {\itshape Racism}, {\itshape Threat and Violence}, {\itshape Personal Insult} among others. It is also noteworthy that classes detected by the {\itshape Radicalization Narratives} model include {\itshape Legitimacy of ideology}, {\itshape Group’s achievements}, {\itshape Promote group ideology} and {\itshape Homophily} that clearly promote partisan narratives and perspectives, omitting discussion, complexity, and diversity of thought.

Table \ref{table:4} shows instead the output of {\itshape Sentiment analysis} and {\itshape Writeprint} models. On average, the information extracted from the datasets undeniably highlights the predominantly negative sentiment. As confirmed by other studies \cite{tsugawa2017relation, hansen2011good, vosoughi2018spread}, it seems a common pattern in the spread of misinformation where negative polarity has been demonstrated to be more related to virality.
Concerning the writeprint indexes, when values are below 14 for the {\itshape Coleman-Liau} \cite{coleman1975computer} and {\itshape Automated Readability} \cite{smith1967automated} indexes, it typically indicates that the texts are easy to process. Conversely, values exceeding 40 for the {\itshape Gulpease index} \cite{lucisano1988gulpease}, often denote highly complex texts. Consequently, all the writeprint index values suggest that the texts are relatively easy for the audience to read and comprehend, contributing to the content's virality.

The results of the analysis, conducted subsequent to the data characterization, reveal that the attributes within these datasets align with research pertaining to misinformation. However, it is important to acknowledge that these attributes might not suffice for the development of an effective classifier for misleading content. Therefore, our next step involves evaluating the predictive capacity of these features for the study's tasks.

\subsection{Predictiveness of Symbolic Features}
To explore the question raised in RQ2 about the validity of symbolic resources, we employ the feature selection method outlined in section 4. This method aids us in evaluating the predictiveness of information captured by symbolic models discussed in section 3.2. 
For each specific task, Table \ref{table:5} shows the top 3 features selected by each symbolic model. The tables reveal that the smaller the number of samples per task, the smaller the number of predictive features, and this is especially true when the scarcity of samples is associated with smaller types of text such as tweets or headlines.
While the most frequent features in Table \ref{table:3} may be indicative of misinformation, they are not always predictive in classification. For example, the {\itshape hatred} and {\itshape racism} classes, preserve the ability to be distinctive across most of the tasks, whereas, distinct features beyond the most commonly occurring classes, such as {\itshape Religious Hatred} and {\itshape Discredit enemy}, unveil stronger association with the target variables.
Tasks with fewer data samples tend to exhibit stronger associations with basic stylometric traits such as {\itshape charactersPerSentence} and {\itshape emoticonsPerSentence}, whereas larger datasets exhibit more pronounced connections with complex readability indexes.

Feature selection analysis revealed that the majority of the tasks can be effectively represented by attributes in line with social science studies discussed in section 2.1. Nevertheless, the most predictive features do not necessarily align with the most frequent ones and the next crucial step is to assess their effectiveness in the implementation of misinformation classifiers.

\begin{table*}
  \Description{Macro accuracy and F1 score values over test sets of misinformation tasks using full training set capacity. On average, the proposed hybrid model AdapterF achieves the highest level of performance.}
  \caption{Macro accuracy and F1 score values over test sets of misinformation tasks using full training set capacity.}
  \begin{tabular}{lcccccccc}
    \toprule
    \multicolumn{1}{c}{} &
    \multicolumn{4}{c}{Baseline} &
    \multicolumn{4}{c}{Ours} \\
    \cmidrule(lr){2-5}
    \cmidrule(lr){6-9}
    \multicolumn{1}{c}{} &
    \multicolumn{2}{c}{RoBERTa} &
    \multicolumn{2}{c}{UNIFIEDM2} &
    \multicolumn{2}{c}{AdapterV} &
    \multicolumn{2}{c}{AdapterF}  \\
    \cmidrule(lr){2-3}
    \cmidrule(lr){4-5}
    \cmidrule(lr){6-7}
    \cmidrule(lr){8-9}
    \multicolumn{1}{l}{Task} &
    \multicolumn{1}{c}{Acc} &
    \multicolumn{1}{c}{F1} &
    \multicolumn{1}{c}{Acc} &
    \multicolumn{1}{c}{F1} &
    \multicolumn{1}{c}{Acc} &
    \multicolumn{1}{c}{F1} &
    \multicolumn{1}{c}{Acc} &
    \multicolumn{1}{c}{F1} \\
    \midrule
    NewsBias & 72.8\% & 65.5\% & 81.0\% & \textbf{70.2\%} & \textbf{81.9\%} & 69.8\% & \textbf{81.9\%} & 69.4\% \\
    FakeNews & 84.3\% & 74.9\% & \textbf{85.4}\% & 73.9\% & 85.0\% & 74.4\% & 85.0\% & \textbf{75.2\%} \\
    Rumor & 87.6\% & 86.9\% & 92.9\% & 92.5\% & 92.4\% & 91.5\% & \textbf{93.6\%} & \textbf{93.0\%} \\
    Clickbait & 84.4\% & 77.4\% & \textbf{86.3\%} & 78.7\% & 85.3\% & 77.9\% & \textbf{86.3\%} & \textbf{80.5\%} \\
    \midrule
    Average & 82.2\% & 76.1\% & 86.4\% & 78.8\% & 86.1\% & 78.4\% & \textbf{86.7\%} & \textbf{79.5\%} \\
    \bottomrule
  \label{table:6}
  \end{tabular}
\end{table*}

\begin{table*}
  \Description{Macro F1 score values over misinformation test sets in fewshot training scenario. On average, the proposed hybrid model AdapterF achieves the highest level of performance in this scenario too.}
  \caption{Macro F1 score values over misinformation test sets in fewshot training scenario.}
  \resizebox{\textwidth}{!}{\begin{tabular}{lcccccccccc}
    \toprule
    \multicolumn{1}{c}{} &
    \multicolumn{3}{c}{10 examples} &
    \multicolumn{3}{c}{25 examples} &
    \multicolumn{3}{c}{50 examples}  \\
    \cmidrule(lr){2-4}
    \cmidrule(lr){5-7}
    \cmidrule(lr){8-10}
    \multicolumn{1}{c}{} &
    \multicolumn{1}{c}{Baseline} &
    \multicolumn{2}{c}{Ours} &
    \multicolumn{1}{c}{Baseline} &
    \multicolumn{2}{c}{Ours} &
    \multicolumn{1}{c}{Baseline} &
    \multicolumn{2}{c}{Ours} \\
    \cmidrule(lr){2-2}
    \cmidrule(lr){3-4}
    \cmidrule(lr){5-5}
    \cmidrule(lr){6-7}  
    \cmidrule(lr){8-8}
    \cmidrule(lr){9-10}  
    \multicolumn{1}{l}{Task} &
    \multicolumn{1}{c}{UNIFIEDM2} &
    \multicolumn{1}{c}{AdapterV} &
    \multicolumn{1}{c}{AdapterF} &
    \multicolumn{1}{c}{UNIFIEDM2} &
    \multicolumn{1}{c}{AdapterV} &
    \multicolumn{1}{c}{AdapterF} &
    \multicolumn{1}{c}{UNIFIEDM2} &
    \multicolumn{1}{c}{AdapterV} &
    \multicolumn{1}{c}{AdapterF} \\
    \midrule
    Propaganda & 56.1\% & 38.4\% & \textbf{60.2\%} & 62.5\% & 50.9\% & \textbf{63.3\%} & \textbf{72.9\%} & 63.5\% & 68.6\% \\
    Fake News Article & 42,4\% & 69.9\% & \textbf{71.0\%} & 53.1\% & 76.3\% & \textbf{82.0\%} & 74.2\% & \textbf{78.4\%} & 74.9\% \\
    Fake News Title & \textbf{55.3\%} & 33.6\% & 35.4\% & 67.0\% & \textbf{77.5\%} & 74.9\% & 71.4\% & \textbf{77.8\%} & 76.8\%\\
    Covid Check-worthy Twitter & 61.7\% & 67.6\% & \textbf{68.0\%} & 64.4\% & 58.2\% & \textbf{70\%} & 73.2\% & 70.0\% & \textbf{74.9\%} \\
    Covid False Twitter Claim & \textbf{54.2\%} & 45.6\% & 45.6\% & \textbf{56.3\%} & 45.9\% & 45.9\% & 59.7\% & 45.9\% & \textbf{71.5\%} \\
    \midrule
    Average & 53.9\% & 51.0\% & \textbf{56.0\%} & 60.6\% & 61.7\% & \textbf{67.2\%} & 70.2\% & 67.1\% & \textbf{73.3\%} \\
    \bottomrule
  \label{table:7}
  \end{tabular}}
\end{table*}

\subsection{Content-based Misinformation Classification}

In this section, we address the question presented in RQ3 by implementing a content-based misinformation classifier enhanced with symbolic features that exhibit some predictability across misinformation tasks and evaluate their effectiveness considering the heterogeneity of this phenomenon.

\subsubsection{Baseline}
We opted for the two RoBERTa large models presented in \cite{lee-etal-2021-unifying}. The first underwent task-specific fine-tuning for each misinformation typology. The second, the {\itshape UnifiedM2} model, is trained in a multi-task learning environment aiming to unify different misinformation types and build richer representations. This last model achieved state-of-the-art results across all tasks.

\subsubsection{Experimental Settings}
The experimentation is conducted on a server equipped with 32GB of RAM and a single NVIDIA GeForce GTX 1080 Ti GPU. Adapter models are implemented using adapter-transformers library\footnote{\url{https://adapterhub.ml/}}, trained for 30 epochs with Adam optimizer \cite{kingma2014adam}. We set a learning rate of 1e-4, early stopping patience to 10, a maximum sequence length of 128, a batch size of 32, a dropout of 0.1, and a layer normalization eps of 1e-12, added for numerical stability, following RobERTa's standard hyperparameters.

\subsubsection{Results}
We replicate the experiments presented in \cite{lee-etal-2021-unifying} by training the model we propose in two distinct manners. The first involves the fine-tuning of task-specific adapters on each dataset presented in Table \ref{table:1}, repeated 3 times with different seeds, using 10\% of the data for development and 15\% of the data for testing. The second manner concerns assessing the generalizability of models in a few-shot learning scenario, where each dataset reported in Table \ref{table:2}, is trained using 10, 25, and 50 samples which we further divide into 80\% training and 20\% validation, employing the remaining part for testing. With {\itshape AdapterV}, we refer to a RoBERTa Pfeiffer adapter with a vanilla classification head, whereas, with {\itshape AdapterF}, we refer to our proposed model, or rather, a RoBERTa Pfeiffer adapter with a custom head in which we integrate features from symbolic models.

{\itshape Fine-tuning}.
We report the findings of the first part of the experiments in Table \ref{table:6}. As highlighted in the table, our {\itshape AdapterF} outperforms other models in terms of both macro F1 and accuracy across the four diverse dataset/task. These results suggest the ability of symbolic models to integrate crucial knowledge for misinformation classification and overcome information that can be acquired by multi-task training. These results prove the predictive power of layered linguistic features across misinformation heterogeneity. Furthermore, our method alleviates the time and resource-intensive process of acquiring additional resources for domain shifting.

{\itshape Few-shot.} To assess our approach's generalizability, we conduct few-shot learning experiments with unknown topics, the results of which are reported in Table \ref{table:7}. Once more, comparing our proposed model with the baseline showcases its consistent superiority across the selected metric. These results underscore the model's strength and competitiveness for state-of-the-art content-based misinformation detection across various domains and resource constraints.

\section{Discussion}

This research work is grounded in the literature of social sciences, with the primary objective of defining the phenomenon of misinformation and its inherent characteristics. By drawing insights from both recent and older works, we have effectively characterized this phenomenon and assessed the compatibility of the data employed by the research community to develop automated models for detecting content-based misinformation.

A distinguishing characteristic of this paper is its exploration of the various layers of linguistic analysis that have evolved through years of research in the field of misinformation language. It employs a multi-layered approach that involves the use of tools designed to address these distinct characteristics. For the first time, this work shows the value of effectively infusing pre-existing symbolic knowledge in a language model architecture for content-based misinformation detection, producing models with SotA performance at a fraction of the cost. To accomplish this, we have utilized proprietary tools that are readily available for use, while also remaining receptive to the possibility of replacing these tools with a variety of open resources made available by the research community. These include models for extracting sentiment from texts, dictionaries tailored for detecting emotions, and models for identifying rhetorical structures or stylometric indexes among others. Any symbolic feature vector can be fed into the proposed network architecture. The combined use of these resources has yielded favorable outcomes in our series of experiments, particularly within the context of knowledge transfer. This highlights how the concurrent application of these stratified resources can establish a framework that offers a comprehensive representation of the phenomenon in question.

Regarding the fusion of symbolic and neural models, we have explored diverse methodologies, spanning from simple concatenation to the selective application of task-specific symbolic models via attention systems, as well as the automation of feature selection by the model itself. Among these methodologies, the model introduced in this study has demonstrated its superiority in both results and efficiency, solidifying its position as the preferred model for this research. The process of infusing knowledge into a language model is an intriguing aspect that we intend to further explore in various contexts, employing different existing methodologies.

\section{Conclusion and Future Work}
In this study, we propose harnessing symbolic linguistic resources inspired by insights from social science research to automate the detection of misinformation based on its content. Our experiments leverage a suite of off-the-shelf freely available symbolic models tailored to identify layered linguistic attributes. 
To ensure the effectiveness of our method, we employ feature selection techniques to identify the optimal feature set for each specific dataset. This information is subsequently combined with the capabilities of language models. Our method is validated across a range of datasets, carefully selected and analyzed to represent the heterogeneous misinformation phenomenon. The outcomes of our research illustrate that, in the ever-evolving real-world context where text topics, domains, and structures constantly evolve, embracing structured knowledge offers substantial advantages. This approach significantly enhances accuracy, efficiency, and resource utilization, ultimately achieving state-of-the-art performance levels. Our approach showcases remarkable generalizability, thereby enhancing its robustness against domain shifts. Our future research will explore injecting this extra knowledge into language models, aiming to grant them the adaptability and generalizability seen in symbolic models in this study.

\begin{acks}
Work supported by the European Comission under grant 101022004 – TRACE – as part of the Horizon 2020 research and innovation programme. \includegraphics[scale=0.02]{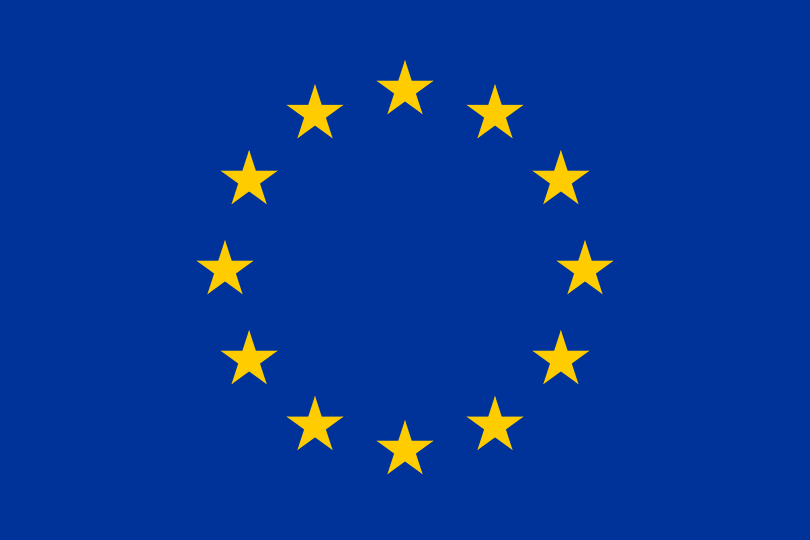}

\end{acks}

\bibliographystyle{ACM-Reference-Format}
\bibliography{sample-base}

\end{document}